\documentclass[letterpaper, 10 pt, conference]{ieeeconf}

\IEEEoverridecommandlockouts
\usepackage[english]{babel}
\usepackage{amsmath}  
\usepackage{amssymb}  
\usepackage{graphicx}
\usepackage{subfigure}
\usepackage{caption}
\usepackage{multirow}
\usepackage{amsfonts}
\usepackage{tabularx}
\usepackage{algorithm,algorithmic}
\usepackage{bm}
\usepackage{breakurl}
\usepackage{enumerate}
\usepackage{adjustbox}
\usepackage{xcolor}
\usepackage{siunitx}
\newcolumntype{b}{X}
\newcolumntype{s}{>{\hsize=.3\hsize}X}
\newcolumntype{x}{>{\hsize=.5\hsize}X}
\usepackage{tikz}
\usetikzlibrary{shapes.geometric, arrows}
\usepackage{array}
\usepackage{cite}
\usepackage{wasysym}
\usepackage{url}
\usepackage{xurl}
\usepackage{booktabs}
\makeatletter
\let\NAT@parse\undefined
\makeatother
\usepackage{hyperref}
\usepackage{cleveref}

\tikzstyle{startstop} = [rectangle, rounded corners, minimum width=3cm, minimum height=1cm,text centered, draw=black, fill=red!30]
\tikzstyle{arrow} = [thick,->,>=stealth]

\title{\LARGE \bf Geometric Fault-Tolerant Control of Quadrotors in Case of\\ Rotor Failures: An Attitude Based Comparative Study}

\author{Jennifer Yeom, Guanrui Li, and Giuseppe Loianno
\thanks{The authors are with the New York University, Tandon School of Engineering, Brooklyn, NY 11201, USA. {\tt\footnotesize email: \{jennifer.yeom, lguanrui, loiannog\}@nyu.edu}.}
\thanks{This work was supported by the DARPA YFA Grant D22AP00156-00, the NSF CAREER Award 2145277, the NSF CPS Grant CNS-2121391, Qualcomm Research, Nokia, and NYU Wireless.}
}

\begin{document}
\maketitle
\thispagestyle{empty}
\pagestyle{empty}

\begin{abstract}
The ability of aerial robots to operate in the presence of failures is crucial in various applications that demand continuous operations, such as surveillance, monitoring, and inspection.
In this paper, we propose a fault-tolerant control strategy for quadrotors that can adapt to single and dual complete rotor failures. Our approach augments a classic geometric tracking controller on $SO(3)\times\mathbb{R}^3$ to accommodate the effects of rotor failures.  We provide an in-depth analysis of several attitude error metrics to identify the most appropriate design choice for fault-tolerant control strategies. To assess the effectiveness of these metrics, we evaluate trajectory tracking accuracies. Simulation results demonstrate the performance of the proposed approach.
\end{abstract}

\section*{Supplementary material}
\textbf{Video}: \url{https://youtu.be/Hp_8VTojDb8}

\IEEEpeerreviewmaketitle

\section{Introduction} \label{sec:introduction}
Quadrotors have gained popularity in various applications in the past couple of decades including, but not limited to search and rescue \cite{almurib2011searchandrescue}, transportation \cite{li2021transportation}, and entertainment such as cinematography and aerial photography \cite{joubert2016cinematographer}. This is mainly because these machines are cost-effective, relatively easy to use, and can hover in place as well as maneuver in constrained and cluttered environments unlike fixed-wing Unmanned Aerial Vehicles (UAVs).

Quadrotors are particularly well-suited for tasks that require maneuverability and stability in a dynamic environment. However, they are under-actuated by design and the failure of one or more rotors is a common occurrence that can severely compromise their operations as well as their stability and performance. As applications of quadrotors expand, so does the importance of ensuring their safety. The inherent possibility of collision, actuator or sensor failures, and external disturbances causing instabilities in a quadrotor's environment necessitates the design of novel algorithms that are resilient to system failures. If a quadrotor is equipped with the ability to detect and recover from a failed rotor, it can continue to perform its mission even in presence of these occurrences. This capability can significantly increase the reliability and safety of quadrotors in real-world applications.

In this work, we explore a full nonlinear control strategy to tackle the fault-tolerant control issue. This work makes two significant contributions. First, we develop of a fault-tolerant control strategy for quadrotors capable of adapting to single and dual complete rotor failures. Our approach involves modifying a standard geometric $SO(3)$ controller to account for the effects of rotors' failures. Second, we analyze how the use of several attitude error metrics impact the algorithm's performance. This allows providing insights into the effectiveness of each method in terms of its ability to accurately track the desired attitude and position, and inform the design of future fault-tolerant control strategies for quadrotors.

\begin{figure}
    \centering
    \includegraphics[width=0.97 \linewidth]{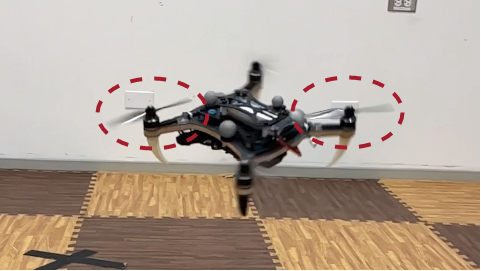}
    \caption{A quadrotor in yawing motion as 2 of its motors are subject to failure. The two failed rotors are circled in red dashed lines. Unlike the healthy rotors that are spinning, the blades are visible.}
    \vspace{-0.5cm}
    \label{fig:intro_image}
\end{figure}

\section{Related Works}~\label{sec:related_works}
Research in the area of fully failed rotors have proposed several ways of ensuring robust control with three or two working rotors. For instance, \cite{freddi2011feedback} develops a feedback linearization approach, which successfully controls a quadrotor with a single rotor failure to follow a trajectory. The work in \cite{freddi2011feedback} is the first to implement a controller that sacrifices the yaw control as the instability for the yaw motion is inevitable after a motor failure. Many studies have followed suit in developing controllers that surrender the control in yaw, only controlling the altitude and attitude in roll and pitch. An emergency landing strategy in case of a rotor failure is implemented in \cite{lippiello2014emergency} and \cite{lippiello2014backstepping} where the researchers design a Proportional Integral Derivative (PID) and backstepping controller for a birotor aerial vehicle, shutting off the opposing faulty rotor. A double rotor failure problem is solved with Sequential Linear Qaudratic (SLQ) control in \cite{deCrousaz2015sql} by continuously linearizing around a predicted trajectory. A sliding mode controller capable of detecting and controlling the fault is presented in \cite{Hou2020NonsingularTS}. All the aforementioned studies are verified only in simulation.

A few works demonstrate real-world flight tests with failed actuators or rotors. For example, \cite{mueller2014lqr} proposes periodic solutions using Linear Quadratic Regulator (LQR) for one, two opposing, and three failed rotor configurations, but the solutions are valid only around linearized points. Furthermore, a hybrid nonlinear controller is designed in \cite{sun2018highspeed} and \cite{sun2021indi} using Incremental Nonlinear Dynamic Inversion (INDI), reaching up to $9$ m/s of flight speed with one and two opposing failed rotors. An event camera and range sensor are used in \cite{sun2021eventcamera} to achieve autonomous flight only using onboard sensors while subjected to a rotor failure. The authors present a state estimation pipeline that enables a quadrotor to hover and track setpoints without the need for a motion capture system. Although the approach offers promising results, the paper focuses on the state estimation problem rather than providing control analysis. One of the latest advancements in this field is the implementation of nonlinear Model Predictive Control (MPC) in conjunction with INDI in \cite{nan2022npmcfault} where a quadrotor with a failed rotor is recovered from dynamic and inverted positions. Similarly to the LQR, this approach relies on Sequential Quadratic Programming and therefore linearizes the dynamics at each iteration. This also implies that although successful in its implementation, the MPC solution involves very high computational loads. In this work we exploit a fully nonlinear properties of geometric controller on SO(3) and quickly solve the fault-tolerant control problem in case of one or multiple rotor failures for this class of approaches.

\section{Methodology} \label{sec:methodology}
\subsection{Preliminaries}
\begin{figure}
    \centering
    \includegraphics[width=0.9\linewidth]{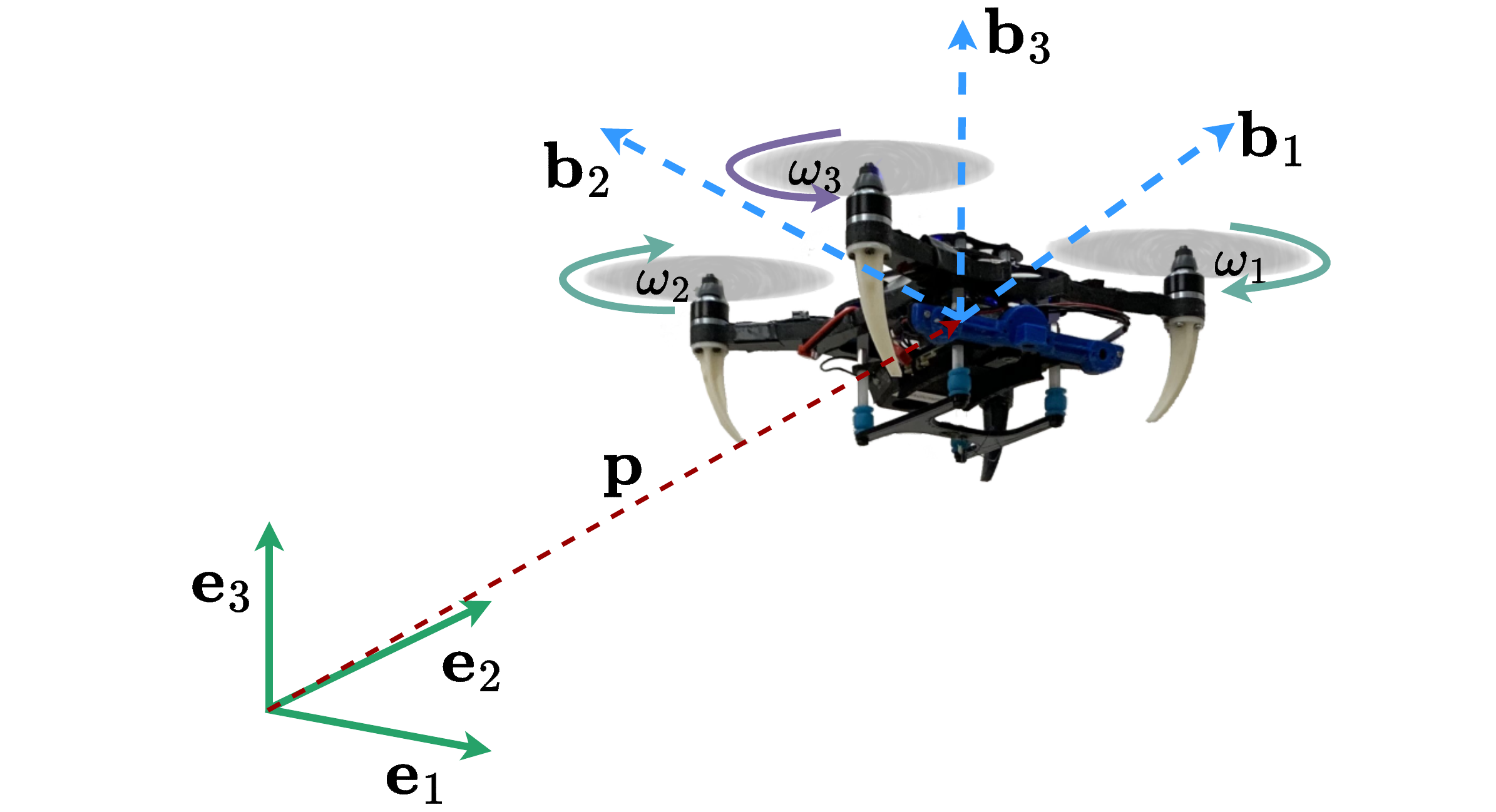}
    \caption{Quadrotor model with inertial and body frame definitions. Rotors 1 and 2 spin clockwise whereas rotors 3 and 4 spin counter-clockwise. $\mathbf{p}$ is the position vector.}
    \label{fig:quadrotor_model}
    \vspace{-0.5cm}
\end{figure}

Consider a quadrotor as shown in Fig.~\ref{fig:quadrotor_model}, we choose the inertial reference frame as $\begin{bmatrix}\mathbf{e}_1& \mathbf{e}_2& \mathbf{e}_3\end{bmatrix}$ and the body fixed frame as $\begin{bmatrix}\mathbf{b}_1& \mathbf{b}_2& \mathbf{b}_3\end{bmatrix}$. The origin of the body frame is aligned with the center of mass of the quadrotor. The $\mathbf{b}_1$ and $\mathbf{b}_2$ axes are defined on the plane where the four rotors lie, $90^\circ$ degrees from each other. The third axis, $\mathbf{b}_3$ points upward from the center of mass and aligns with the thrust vector of the quadrotor. 
The equations of motion of a quadrotor can be written as
\begin{align} \label{eq:1}
\begin{split}
    \dot{\mathbf{p}} &= \mathbf{v},\\
    m \dot{\mathbf{v}} &= m g \mathbf{e}_3 - f \mathbf{R} \mathbf{e}_3 + \mathbf{k}_{td} \mathbf{v},\\
    \dot{\mathbf{R}} &= \mathbf{R} \hat{\mathbf{\Omega}},\\
    \mathbf{M} &= \mathbf{J} \dot{\mathbf{\Omega}} + \mathbf{\Omega} \times \mathbf{J} \mathbf{\Omega} + \mathbf{k}_{rd} \mathbf{\Omega},
\end{split}
\end{align}
where $\mathbf{p} = \begin{bmatrix}x& y& z\end{bmatrix}^\top$, and $\mathbf{v}$ are the position and velocity of the center of mass of the quadrotor. $\mathbf{R}$, $f$, $m$, $g$ are the attitude, thrust, mass of the quadrotor, and gravity respectively, $\mathbf{\Omega}=\begin{bmatrix}p& q& r\end{bmatrix}^\top$ are the angular velocities with respect to the body frame, $\mathbf{M}=\begin{bmatrix} M_1& M_2& M_3\end{bmatrix} $ are the moments in the body frame, and $\mathbf{J}$ is the inertial matrix of the quadrotor. The \textit{hat map}, $\hat{\cdot}$ : $\mathbb{R}^{3} \rightarrow so(3)$  represents the mapping such that $\hat{\mathbf{a}}\mathbf{b} = \mathbf{a} \times \mathbf{b}, \forall \ \mathbf{a},\mathbf{b} \in \mathbb{R}^{3}$.  Lastly, $\mathbf{k}_{td}$ and $\mathbf{k}_{rd}$ are the translational and rotational drag coefficients, respectively. 

It is especially important to model the rotational drag as our solution to fault-tolerant control is a spinning quadrotor. If the rotational drag is not accounted for, the simulated quadrotor will spin at a continuously increasing rate as it approaches physically impossible angular velocities. Thus, we choose to implement a simple linear drag model obeying Stoke's law as in eq.~(\ref{eq:1}). The translational drag is set proportional to the linear velocity and the rotational drag is set proportional to the angular velocity \cite{freddi2011feedback}.

\subsection{Control Design} \label{sec:control_design}

\begin{figure*}
    \centering
    \includegraphics[width=\linewidth]{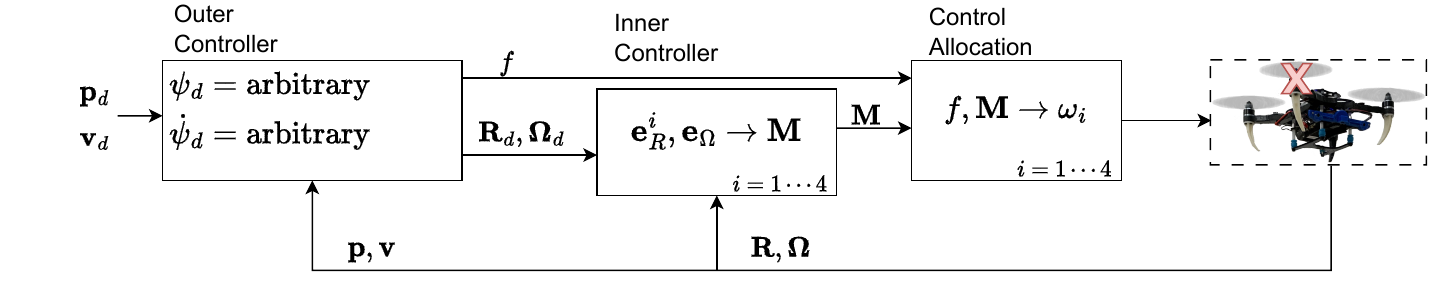}
    \caption{Fault-tolerant controller block diagram.}
    \label{fig:control_diag}
\end{figure*}

We adapt the geometric controller derived in \cite{lee2010so3} for the possibility of losing one or two opposing rotors in flight. As shown in Fig.~\ref{fig:control_diag}, the outer controller solves for the desired attitude ($\mathbf{R}$), thrust ($f$), and angular velocities ($\mathbf{\Omega}$). The inner controller solves for the moments required to control the attitude of the quadrotor. The thrust and moment information is then used to allocate desired motor speed commands to each of the rotors. 

\subsubsection{Outer-loop Controller}
In our approach for fault-tolerant control adaptation of the current state-of-the-art geometric control method, we use the circumstance of losing a motor, which equates to losing a degree of freedom, to our advantage. We choose to give up the control of the yaw as a quadrotor can still remain in a stable and controllable hover condition while spinning in the yaw direction. This situation can be used to our advantage because the rotation matrix or attitude of a quadrotor, $\mathbf{R} = \begin{bmatrix} \mathbf{b}_1& \mathbf{b}_2& \mathbf{b}_3 \end{bmatrix} \in SO(3)$, can be described as the product of the yaw rotation matrix and the tilt rotation matrix, as in
\begin{equation} \label{eq:2}
\mathbf{R} = \mathbf{R}_{\psi} \mathbf{R}_{\phi}, 
\end{equation}
where $\mathbf{R}_{\psi}$ is the rotation matrix of the yaw and $\mathbf{R}_{\phi}$ is the rotation matrix of the tilt. The two matrices are defined as
\begin{equation} \label{eq:R_yaw}
R_{\psi} = 
\begin{bmatrix}
 \cos(\psi)& -\sin(\psi) & 0\\ 
 \sin(\psi)& \cos(\psi) & 0 \\ 
 0& 0 & 1
\end{bmatrix},
\end{equation}
\begin{equation} \label{eq:R_tilt}
R_{\phi} = 
\begin{bmatrix}
b_{3z} + \frac{b_{3y}^2}{1+b_{3z}} & \frac{-b_{3x} b_{3y}}{1 + b_{3z}} & b_{3x}\\ 
\frac{-b_{3x} b_{3y}}{1 + b_{3z}}  & 1 - \frac{b_{3y}}{1+b_{3z}}  & b_{3y}\\ 
-b_{3x} & -b_{3y} & b_{3z}
\end{bmatrix},
\end{equation}
where $\mathbf{b}_3 = \begin{bmatrix}b_{3x}& b_{3y}& b_{3z}\end{bmatrix} ^\top$ is the third column of the quadrotor's orientation, $\mathbf{R}$.
This decomposition is extremely useful because our control of a quadrotor with failed rotors is a solution without the control of the yaw.

In order to adapt the geometric controller to partial or complete rotor failures, we remove the control of the yaw from the controller entirely. As a product, the desired yaw and yaw rate can be set arbitrarily. We set both to $0$ in our implementation of the fault tolerant controller, however the controller does not command the quadrotor to correct to $0$ yaw or $0$ yaw rate.

As per the standard geometric control algorithm, the thrust vector is normalized and chosen as the third body axis, $\mathbf{b}_3$ as in
\begin{equation}\label{eq:b3}
    \mathbf{b}_3 = \frac{f \mathbf{R}\mathbf{e}_3}{\|f \mathbf{R}\mathbf{e}_3 \|}.
\end{equation}
Next, the second column of $\mathbf{R}_\psi$ is used as the desired heading, $\mathbf{b}_{hd}$. Here, the resulting $\mathbf{b}_{hd}$ is set to $\begin{bmatrix}0& 1& 0 \end{bmatrix}^\top$ per the desired yaw and yaw rate. The remaining body axes $\mathbf{b}_1$ and $\mathbf{b}_2$ are defined by
\begin{equation} \label{eq:b1}
\mathbf{b}_1 = \frac{\mathbf{b}_{hd} \times \mathbf{b}_3}{\|\mathbf{b}_{hd} \times \mathbf{b}_3 \|},
\end{equation}
\begin{equation} \label{eq:b2}
\mathbf{b}_2 = \frac{\mathbf{b}_3 \times \mathbf{b}_1}{\|\mathbf{b}_3 \times \mathbf{b}_1 \|}.
\end{equation}
The first control input, thrust, is solved for with
\begin{equation} 
f = m \left(\mathbf{k}_p \mathbf{e}_p + \mathbf{k}_v \mathbf{e}_v + \ddot{\mathbf{p}}_d + g \mathbf{e}_3\right) \cdot \mathbf{R} \mathbf{e}_3,
\label{eq:thrust_command}
\end{equation}
where $\mathbf{k}_p$ and $\mathbf{k}_v$ are positive gains and the errors in position and velocity are defined as
\begin{equation} \label{eq:pos_er}
\mathbf{e}_p = \mathbf{p} - \mathbf{p}_d,
\end{equation}
\begin{equation} \label{eq:vel_er}
\mathbf{e}_v = \mathbf{v} - \mathbf{v}_d.
\end{equation}
Lastly, the angular velocities $\mathbf{\Omega}$, are solved for using the relationship 
\begin{equation} \label{eq:omega_hat}
    \mathbf{\hat{\Omega}} = \mathbf{R}^\top \dot{\mathbf{R}},
\end{equation} 
where the $\dot{\mathbf{R}}$ term is solved for by taking the derivative of the body axes, $\mathbf{b}_1, \mathbf{b}_2$, and $\mathbf{b}_3$.  

\subsubsection{Inner-loop Controller}
The inner controller solves for the second control input, moment, from the error in attitude and error in angular velocity. In a traditional geometric controller \cite{lee2010so3}, the attitude tracking error, $\mathbf{e}_R$ is calculated by using the full rotation matrices. Four different types of attitude error metrics will be presented in the following section. 
The angular velocity error, $\mathbf{e}_\Omega$ is calculated by 
\begin{equation} \label{eq:e_omega}
\mathbf{e}_\Omega = \mathbf{\Omega} - \mathbf{R}^\top \mathbf{R}_d \mathbf{\Omega}_d.
\end{equation}
Finally the moments are calculated by
\begin{equation}
\mathbf{M} = -\mathbf{k}_R \mathbf{e}_R - \mathbf{k}_{\Omega} \mathbf{e}_{\Omega} + \mathbf{\Omega}\times \mathbf{J} \mathbf{\Omega}.
\label{eq:moment_commands}
\end{equation}

\subsection{Attitude Metrics}
Inspired by the investigation of attitude error metrics in \cite{spitzer2020error}, we evaluate four total avenues of attitude error representation for use in our fault-tolerant controller. 
\subsubsection{Full Attitude Metric} 
The first error metric uses the desired and estimated full rotation matrices
\begin{equation} \label{eq:e_1}
    \mathbf{e}^1_R = \frac{1}{2} (\mathbf{R}_d^\top \mathbf{R} - \mathbf{R}^\top \mathbf{R}_d)^{\vee}.
\end{equation}
where $\mathbf{R}_d$ is given from the outer controller and $\mathbf{R}$ is the current estimated attitude. The \textit{vee map} $^\vee$ : $so(3) \rightarrow \mathbb{R}^{3}$ is the reverse hat map. Using Rodrigues' formula, this metric can be rewritten as
\begin{equation} \label{eq:e_1_axisangle}
    \frac{1}{2} (\mathbf{R}_e - \mathbf{R}_e^\top)^{\vee} = \sin \rho_e \mathbf{n}_e,
\end{equation}
where $\rho_e$ is the angle and $\mathbf{n}_e$ is the unit vector of the axis angle representation \cite{siciliano2009kinematics} of the full rotation error metric ($\mathbf{R}_e = \mathbf{R}_d^\top \mathbf{R}$), with the definitions

\begin{equation} \label{eq:axis_angle_rho}
    \mathbf{\rho}_e = \cos^{-1} \frac{\text{trace}(\mathbf{R}_e) - 1}{2},
\end{equation}
\begin{equation} \label{eq:axis_angle_n}
    \mathbf{n}_e = \frac{1}{2 \sin(\mathbf{\rho}_e)} \begin{bmatrix}
r_{32} - r_{23}\\ 
r_{13} - r_{31}\\ 
r_{21} - r_{12}
\end{bmatrix}.
\end{equation} 

\subsubsection{Half Angle Metric}
Next we adapt an attitude error metric defined on the special orthogonal group in \cite{Lee2012ExponentialSO} as 
\begin{equation} \label{eq:e_2}
    \mathbf{e}^2_R = 2 \sin \left( \frac{\rho_e}{2} \right) \mathbf{n}_e.
\end{equation}
This metric has a property that the magnitude of the control input is proportional to the given attitude error \cite{Lee2012ExponentialSO}. This approach addresses the fact that the response of the original error metric in $\mathbf{e}^1_R$ weakens as error increases and becomes stronger as the error approaches $90^{\circ}$. This characteristic may be advantageous to the fault-tolerant control as the response of the error metric does not weight as heavily on different degrees of the error.

\subsubsection{$\mathcal{S}^{2}$ Metric}
For the rest of the error metrics, we specifically consider metrics that decompose attitude into yaw and tilt, commonly known as the `reduced attitude control' \cite{chaturvedi2011attitudecontrol}. Readers are reminded of the relationship in eq.~(\ref{eq:2}) for the decomposition of the attitude of a quadrotor into yaw and tilt. First, we define the error in the tilt orientation or rotation matrix as 
\begin{equation} \label{eq:R_tilt_er}
    \mathbf{R}_{\phi_e} = \mathbf{R}^\top_{\phi_{des}} \mathbf{R}_\phi.
\end{equation}
The error in the tilt can also be represented in the axis angle $\mathbf{\rho}_{\phi_e}$ and the corresponding unit vector, $\mathbf{n}_{\phi_e}$. The error in yaw can similarly be represented in axis angle form of $\mathbf{\rho}_{\psi_e}$ and $\mathbf{n}_{\psi_e}$.
An error metric used in \cite{kooijman2019s2s1} for attitude control decomposed by tilt and yaw is
\begin{equation} \label{eq:e_3}
\mathbf{e}^3_R  =
    \begin{cases}
      \sin \rho_{\phi_e} \mathbf{n}_{\phi_e} & \rho_{\phi_e} \leq \frac{\pi}{2}\\
      \mathbf{n}_{\phi_e} & \frac{\pi}{2} < \rho_{\phi_e} < \pi\\
    \end{cases}.
\end{equation}
This error metric is specifically for control around the $\mathbf{b}_1$ and $\mathbf{b}_2$ axes, computing error only in $\mathcal{S}^{2}$. An independent yaw controller is also presented in \cite{kooijman2019s2s1} but omitted in this paper as our fault-tolerant controller does not attempt to control yaw. This error metric in $\mathcal{S}^{2}$ fits particularly well to our problem at hand of only controlling attitude while surrendering control of the yaw angle.

\begin{figure*}[h]
    \centering
    \includegraphics[width=0.9\linewidth, trim=0 550 0 0, clip]{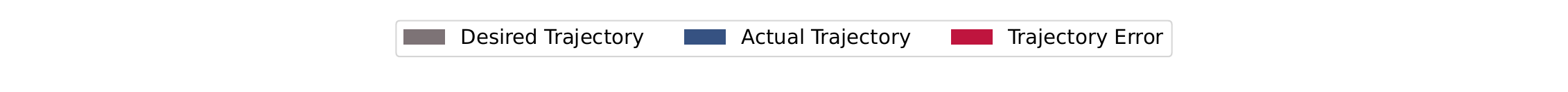}
    \includegraphics[width=1\linewidth, trim=0 0 0 0, clip]{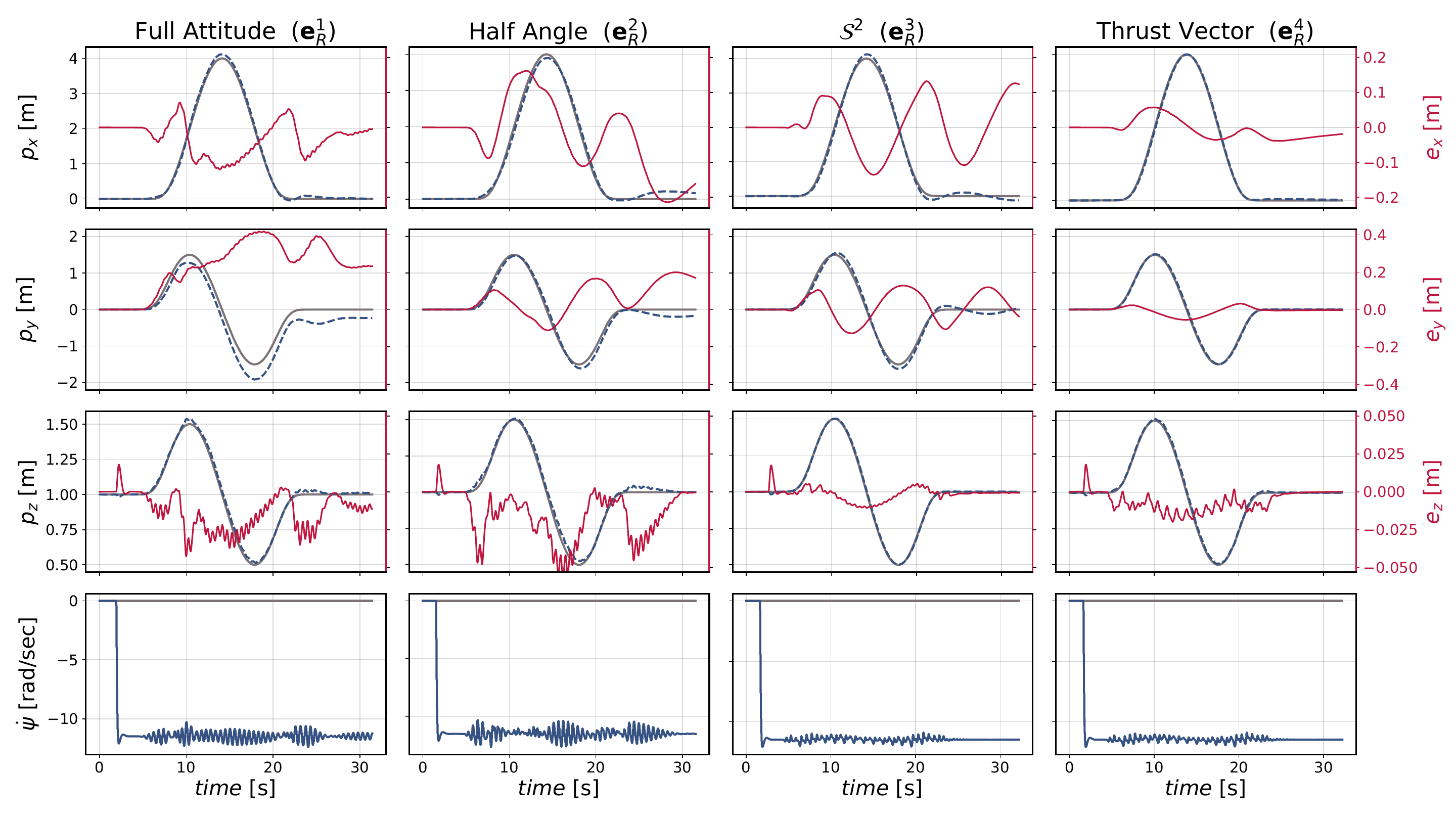}
    \hfill
    \vspace{-2em}
    \caption{Circular trajectory ($2\mathrm{m} \times 1.5\mathrm{m} \times 0.5\mathrm{m}$, 15 sec) performance by $x, y, z$ axes and yaw rate for 1 failed rotor case. Each column is the result of a different error metric. The desired (gray) and actual trajectory (blue, dashed) are overlaid with the error (red) in positions $x, y,$ and $x$. The errors are represented in a separate $y$-axis, depicted on the right of the figures in red. The last row of figures show the desired and actual yaw rate.}
    \label{fig:circ_1}
\end{figure*}

\subsubsection{Thrust Vector Metric}
We define the unit vector that aligns with the thrust vector $\mathbf{n}$, 
\begin{equation} \label{eq:n}
    \mathbf{n} = \mathbf{R} \mathbf{e}_3,
\end{equation}
which is the third vector of a quadrotor's rotation matrix with respect to the world frame. 

Then, we can define the last error metric to evaluate as
\begin{equation} \label{eq:e_4}
    \mathbf{e}^4_R = \mathbf{n}_d \times \mathbf{n},
\end{equation}
where $\mathbf{n}_d$ is the unit vector along the desired thrust vector from the outer controller and $\mathbf{n}$ is the estimated current thrust vector. As the two unit vectors ($\mathbf{n}_d$ and $\mathbf{n}$) represent desired and estimated directions, the magnitude of the cross product is used as a measure of how far off the actual direction is from the desired direction. The cross-product is used as the direction of correction needed to bring the actual direction aligned with the desired direction. 

\subsection{Control Allocation}
The mapping between the motor speeds to the thrust and moments are shown as
\begin{equation} \label{eq:full_allocation}
\begin{bmatrix}
f\\ 
M_1\\ 
M_2\\ 
M_3
\end{bmatrix}
 = 
\begin{bmatrix}
k_f & k_f & k_f & k_f\\ 
0 & 0 & d k_f & -d k_f\\ 
-d k_f & d k_f & 0 & 0\\ 
k_m & k_m & -k_m & -k_m
\end{bmatrix}
\begin{bmatrix}
\omega_1^2\\ 
\omega_2^2\\ 
\omega_3^2\\ 
\omega_4^2
\end{bmatrix}.
\end{equation}
After we obtain the thrust and moments from eq.~\eqref{eq:thrust_command} and eq.~\eqref{eq:moment_commands}, we allocate them to 
motor speed commands of each of the rotors by inverting the above mapping.

For a single rotor failure case, the last row of the allocation matrix and the column corresponding to the failed rotor are removed, resulting in a $3\times3$ matrix. The removal of the last row is significant as this disconnects the control law from all yaw moments, surrendering the control in yaw entirely. Without the loss of generality, the allocation relationship when rotor 1 fails becomes 
\begin{equation} \label{eq:3_allocation}
    \begin{bmatrix}
f\\ 
M_1\\ 
M_2\\ 
\end{bmatrix}
 = 
\begin{bmatrix}
k_f & k_f & k_f\\ 
0 & d k_f & -d k_f\\ 
d k_f & 0 & 0\\ 
\end{bmatrix}
\begin{bmatrix}

\omega_2^2\\ 
\omega_3^2\\ 
\omega_4^2
\end{bmatrix}.
\end{equation}
In the case of two failed opposing rotors, the last row and two columns corresponding to the failed rotor are removed, resulting in a $2\times 2$ matrix. For example, if rotors 1 and 2 fail, the resulting allocation matrix is 
\begin{equation} \label{eq:2_allocation}
    \begin{bmatrix}
f\\ 
M_1\\ 
\end{bmatrix}
 = 
\begin{bmatrix}
k_f & k_f\\ 
d k_f & -d k_f\\ 
\end{bmatrix}
\begin{bmatrix}
\omega_3^2\\ 
\omega_4^2
\end{bmatrix}.
\end{equation}
As both the $3\times 3$ and $2\times 2$ matrices remain full rank, they are always invertible allowing a solution for control allocation despite rotor failures. Of note, the single rotor failure case uses two moments (roll and pitch) whereas the dual rotor failure case uses a single moment to control its attitude. The spinning dynamic of the failed quadrotor makes it possible for the dual rotor failure case to maneuver in the Cartesian space with only a single moment.

\begin{figure*}[t!]
    \centering
    \includegraphics[width=0.9\linewidth, trim=0 550 0 0, clip]{figures/legend_2d.pdf}
    \includegraphics[width=1\linewidth, trim=0 0 0 0, clip]{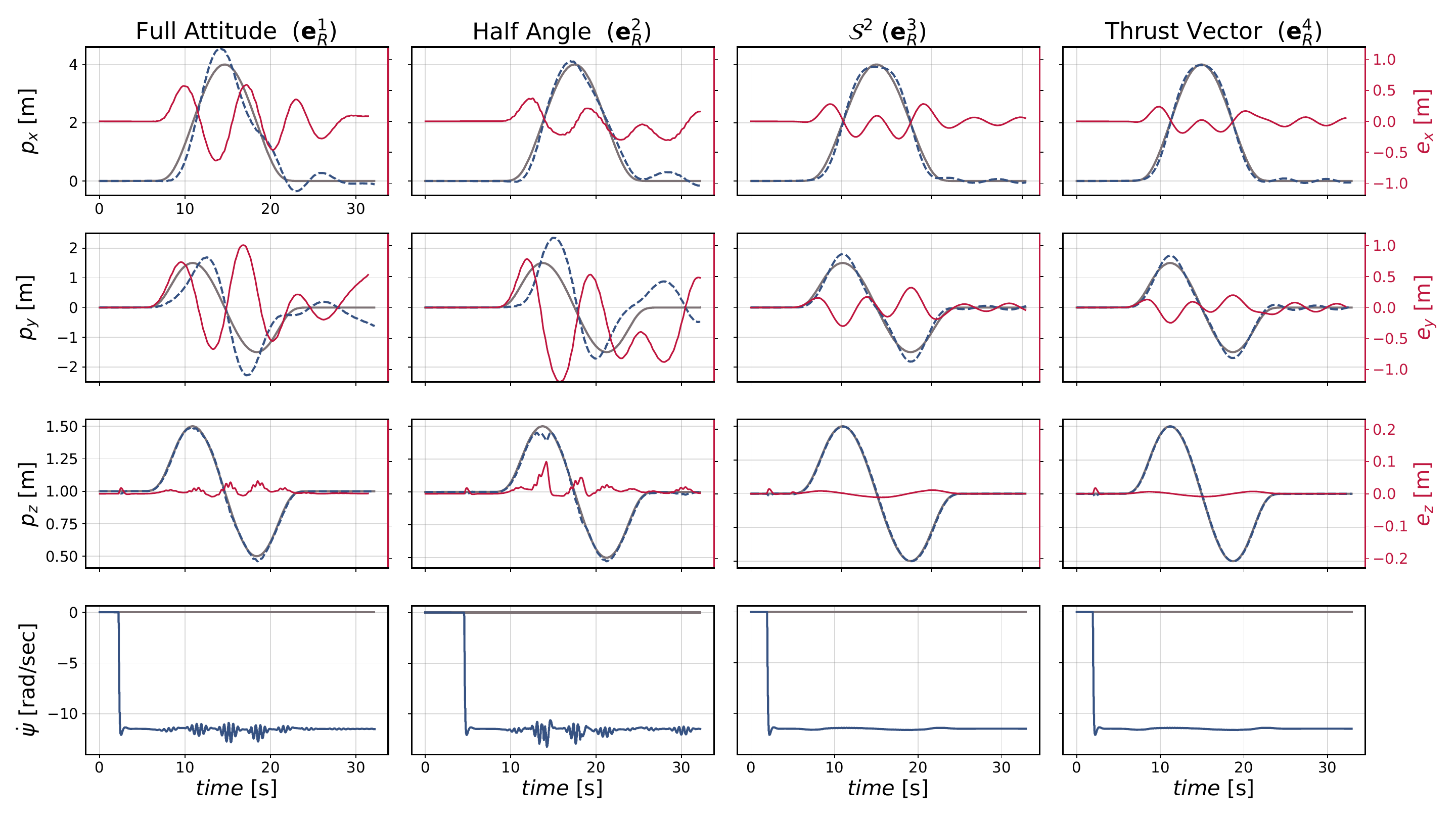}
    \hfill
    \vspace{-2em}
    \caption{Circular trajectory ($2\mathrm{m} \times 1.5\mathrm{m} \times 0.5\mathrm{m}$, 15 sec) performance by $x, y, z$ axes and yaw rate for 2 failed rotors case. Each column is the result of a different error metric. The desired (gray) and actual trajectory (blue, dashed) are overlaid with the error (red) in positions $x, y,$ and $x$. The errors are represented in a separate $y$-axis, depicted on the right of the figures in red. The last row of figures show the desired and actual yaw rate.}
    \label{fig:circ_2}
    \vspace{-1.5em}
\end{figure*}

\section{Results} \label{sec:results}
All simulation experiments are completed in the ROS environment with system dynamics and parameters of a Dragonfly quadrotor~\cite{LoiannoRAL2017}. A rotor fault is injected in the simulation, shutting off one or two opposing rotors while the quadrotor is airborne. We evaluate the fault-tolerant controller's behavior while tracking a trajectory post rotor fault for all four attitude error metrics discussed in Section~\ref{sec:methodology}. 

\subsection{Single Rotor Failure}
After the failure of 1 rotor is injected, the quadrotor is given a command to move in an oval of radius $2$~m in $x$, $1.5$~m in $y$, and $0.5$~m in $z$. This trajectory showcases the quadrotor's ability to maneuver in the cartesian space $x, y, z$ even with a faulty rotor. Fig.~\ref{fig:circ_1} shows the overall tracking performance of the quadrotor for all four attitude representations in $x, y, z$ as well as yaw rate, $\dot{\psi}$. As designed, the desired yaw rate is constant at 0, however it can be seen the quadrotor spins at approximately $-12$ rad/sec as the controller does not regulate the yaw or yaw rate. This yaw rate is a consequence of the angular drag relationship in eq.~(\ref{eq:1}) which is heavily dependent on the $\mathbf{k}_{rd}$ value.

In addition to the first trajectory, we analyze three more circular trajectories with varying speeds and accelerations. We vary the ramp speed and completion time of each trajectory in simulation. For each case, we calculate and report the RMSE separately for all three axes in Table~\ref{tab:rmse1}. The last column of the table shows the RMSE for the controller using the first error metric, $\mathbf{e}^1_R$, with four healthy rotors as a baseline performance. As anticipated, errors generally increase as the same trajectory is tracked at faster speeds which leads to greater roll and pitch angles in the quadrotor's dynamics. However, we cannot report the RMSE for the first and second error metrics in the 5 second trajectory because the controller fails to complete the desired trajectory using these metrics, resulting in a crash. Data analysis reveals that the desired moments in roll and pitch spike when the angles exceed approximately 65$^\circ$, driving the quadrotor to an uncontrollable state. Nevertheless, our results demonstrate that the third and fourth metrics achieve similar errors to a quadrotor with no failures, even at higher speeds around $2.4$~m/s.
\begin{table}[t!]
\centering
\caption {Tracking RMSE (in meters) for one rotor failure case for four different trajectory speeds.}
\begin{tabular}{ccccccc}
\toprule\toprule
duration [sec] & \multirow{2}{*}{axis}  & \multirow{2}{*}{$\mathbf{e}^1_R$} & \multirow{2}{*}{$\mathbf{e}^2_R$} & \multirow{2}{*}{$\mathbf{e}^3_R$} & \multirow{2}{*}{$\mathbf{e}^4_R$} & \multirow{2}{*}{no fault}\\ 
$(\bm{v}_{\max}~[\SI{}{\meter/\second}])$ & & & & & &\\
\midrule\midrule
\multirow{3}{*}{\shortstack{15\\(0.8)}}& $x$  & 0.111   &  0.114   &   0.079    &    0.028  & 0.027    \\
& $y$  & 0.291   &   0.107  &   0.074    &   0.021   & 0.014   \\
 & $z$  & 0.025   &   0.021   &   0.005    &   0.008   & 0.004   \\ \hline
\multirow{3}{*}{\shortstack{12\\(1.0)}}& $x$  & 0.117   &   0.111   &   0.091    &   0.051   &  0.022  \\
& $y$  &  0.344  &   0.079   &   0.089    &    0.046  & 0.014   \\
& $z$  &  0.030   &   0.029   &    0.006  &   0.006   &  0.005 \\ \hline
\multirow{3}{*}{\shortstack{8\\(1.5)}}& $x$  & 0.096    &  0.131    &   0.093    &  0.078   &  0.038  \\
& $y$  & 0.877    &   0.538   &   0.087    &  0.079   &   0.024  \\
& $z$   & 0.045   &   0.059   &   0.012    &   0.015  &  0.007  \\ \hline
\multirow{3}{*}{\shortstack{5\\(2.4)}}& $x$   &  -      &   -      &   0.132    &   0.151   &  0.149 \\
& $y$  &  -       &    -     &   0.075    &  0.089   &   0.103  \\
& $z$   &  -      &   -      &    0.043 &    0.042   &  0.032  \\ 
\bottomrule\bottomrule
\end{tabular}
\label{tab:rmse1}
\end{table}

Overall, it is apparent that the third and fourth attitude error metrics perform much better than the other two metrics. The full error metric produces overshoots and undershoots as well as an offset at the final setpoint of the trajectory. The second error metric exhibits better tracking performance but has offsets and slight oscillations at the final setpoint. This can be seen from the oscillating error in $x$ and $y$ in Fig. \ref{fig:circ_1}. The third and fourth error metrics outperform the rest as they are specifically designed to separate and disregard yaw information when closing the loop of the control logic. 

The first $15$ second circular trajectory is depicted in $3$D space in Fig.~\ref{fig:3d}, showing a more intuitive and side by side comparison of each of the trajectories. Again, the first error metric using the full attitude information is seen to perform worse than the others.


\subsection{Two Rotors Failure}
\begin{figure*}[t!]
    \centering
    \includegraphics[width=0.9\linewidth, trim=0 550 0 0, clip]{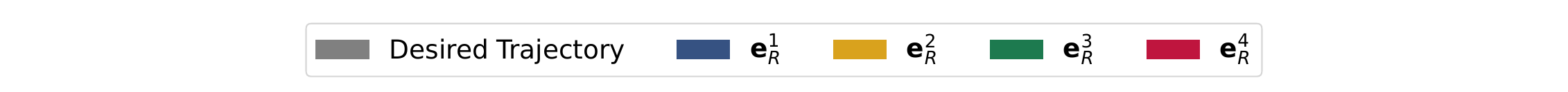}
    \includegraphics[width=1\linewidth, trim=0 50 0 0, clip]{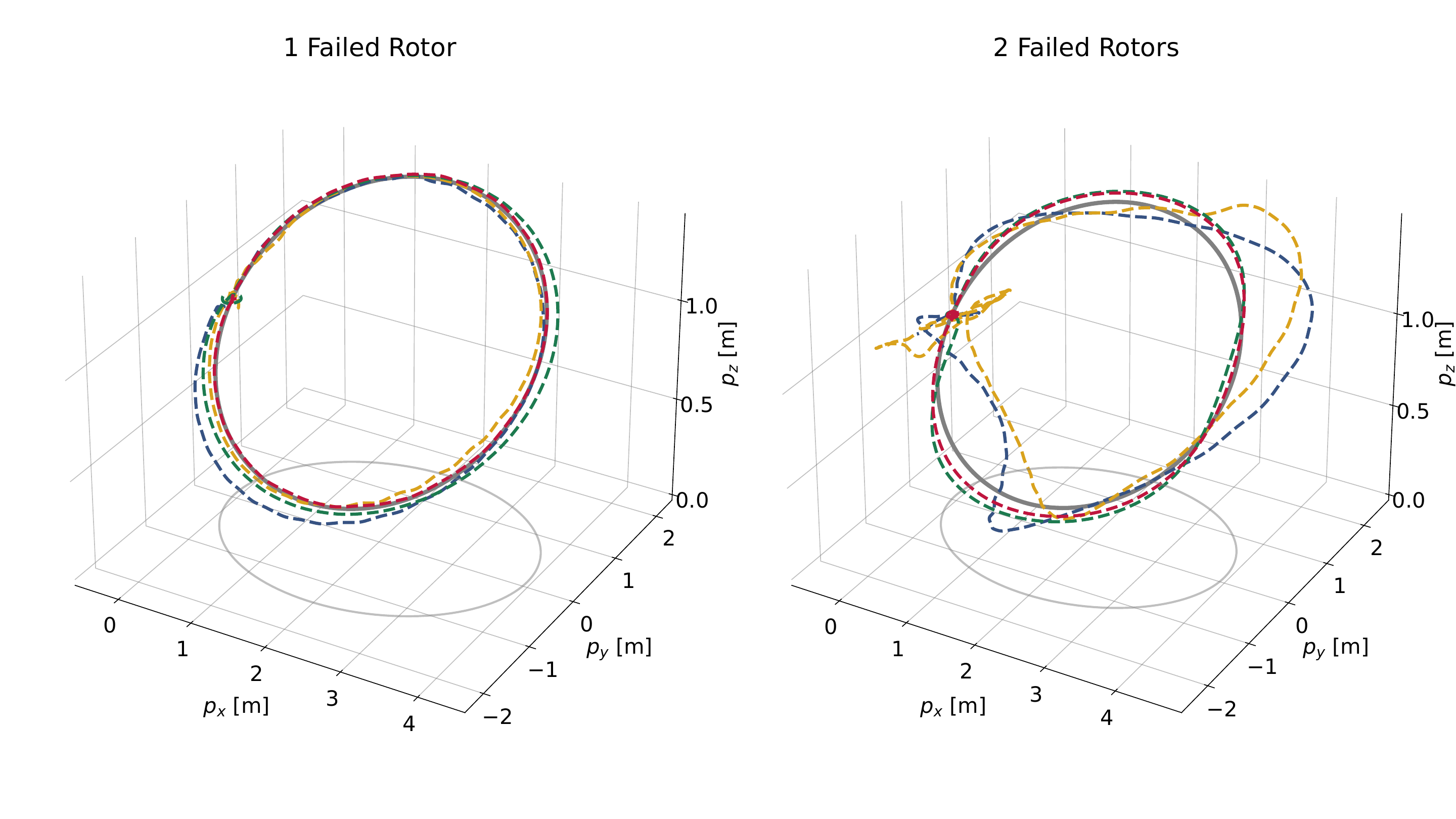}
    \hfill
    \vspace{-1em}
    \caption{3D circular trajectory tracking trace for 1 failed (left) and 2 failed (right) rotors. Performance for 1 failed rotor is visibly better than the 2 failed rotor case. The third and fourth error metrics clearly outperform the first two error metrics.}
    \label{fig:3d}
    \vspace{-1.5em}
\end{figure*}

The same experiments are completed for the $2$ failed rotor case. The tracking performance in $x, y,$ and $z$ are shown in Fig.~\ref{fig:circ_2} and the RMSE for variable trajectories are reported in Table~\ref{tab:rmse2}. The first two metrics result in large deviations from the desired setpoint at the end of the trajectory. The large departure from the reference trajectory can also be seen in the $3$D plot in Fig.~\ref{fig:3d}. Overall, there are larger and more frequent overshoots and course corrections when attempting to follow the same circular trajectory compared to the $1$ failed rotor scenario. 

Tracking performance is both quantitatively and qualitatively worse than the $1$ failed rotor case as the quadrotor is only able to use a single moment to control its attitude. The comparison of error metrics reveals that, in the scenario of two rotor failures, the $\mathcal{S}^{2}$ and thrust vector metrics outperform the full attitude and half angle metrics. For higher speed trajectories, the $\mathcal{S}^{2}$ metric has consistently lower RMSE than the thrust vector metric. Notably, the RMSE values for the $\mathbf{e}^1_R$ and $\mathbf{e}^2_R$ error metrics for the 8 and 5 second trajectories are not included in Table~\ref{tab:rmse2} due to the controller's inability to complete the trajectories using these error metrics. As with the one failed rotor case, the calculated moment diverges when the pitch or roll angles exceed approximately $65^\circ$, preventing the quadrotor from reaching the desired attitude. Although there are apparent divergence from the desired trajectory, we have demonstrated that even with failure of two rotors, the quadrotor is able to maneuver albeit slowly in the Cartesian plane to a desired location for a safe landing. 
\begin{table}[t!]
\centering
\caption {Tracking RMSE (in meters) for the two rotors failure case for four different trajectory speeds.\label{tab:rmse2}}
\begin{tabular}{ccccccc}
\toprule\toprule
duration [sec] & \multirow{2}{*}{axis}  & \multirow{2}{*}{$\mathbf{e}^1_R$} & \multirow{2}{*}{$\mathbf{e}^2_R$} & \multirow{2}{*}{$\mathbf{e}^3_R$} & \multirow{2}{*}{$\mathbf{e}^4_R$} & \multirow{2}{*}{no fault}\\ 
$(\bm{v}_{\max}~[\SI{}{\meter/\second}])$ & & & & & &\\
\midrule\midrule
\multirow{3}{*}{\shortstack{15\\(0.8)}}&$x$  & 0.298 & 0.195  &  0.128  &  0.124  & 0.027    \\
&$y$  & 0.451  &  0.761 &  0.126 &   0.122  & 0.014    \\
&$z$  & 0.015  &  0.035 &  0.005  &  0.004  & 0.004     \\ 
\hline
\multirow{3}{*}{\shortstack{12\\(1.0)}}&$x$  & 0.464 & 0.184  &  0.161  &  0.425  & 0.022    \\
&$y$  & 0.629  &  0.581 &  0.156 &   0.401  & 0.014    \\
&$z$  & 0.022  &  0.029 &  0.005  &  0.005  & 0.005     \\ \hline
\multirow{3}{*}{\shortstack{8\\(1.5)}}&$x$  & - & -  &  0.368  &  0.676 & 0.038    \\
 &$y$  & - &  - &  0.350 &   0.634  & 0.024    \\
&$z$  & - &  - &  0.011  &  0.006  & 0.007     \\ \hline
\multirow{3}{*}{\shortstack{5\\(2.4)}}&$x$  & - & -  &  0.523  &  0.802  & 0.149    \\
&$y$  & - &  - &  0.567 &   0.735  & 0.103    \\
&$z$  & -  &  - &  0.024  &  0.013  & 0.032     \\ 
\bottomrule\bottomrule
\end{tabular}
\vspace{-10pt}
\end{table}

\section{Conclusion} \label{sec:conclusion}
This paper presented a study of different attitude error metrics for the implementation in a fault-tolerant control scheme. Our analysis demonstrates that using rotational error metrics that decompose attitude error into tilt and yaw components leads to better control performance in fault-tolerant control compared to metrics that employ the full attitude error. Metrics using only $\mathcal{S}^{2}$ or the thrust vector measured far better than the other metrics as they were designed to implement a reduced attitude control without attempting to control the yaw. However, we also see merit in using the full attitude and yaw information to avoid redesigning an already existing control law, as the quadrotor is still controllable at lower speed regimes using the more traditional attitude error metrics. 

Our future work will include implementation of this fault-tolerant control scheme to an actual quadrotor for flight tests in the real world. Additional directions for continuation of this research will be in the fault detection and diagnosis realm. We will attempt to use estimation techniques to detect the loss of a rotor's effectiveness, investigate the existence of additional allocation techniques, and the impact of the proposed approaches on multi-robot collaborative tasks. 

\bibliographystyle{IEEEtran}
\bibliography{references}
\end{document}